\documentclass[conference]{IEEEtran}
\IEEEoverridecommandlockouts
\usepackage{cite}
\usepackage{amsmath,amssymb,amsfonts}
\usepackage{algorithmic}
\usepackage{graphicx}
\usepackage{multirow}
\usepackage{textcomp}
\usepackage{balance}
\usepackage{xcolor}
\def\BibTeX{{\rm B\kern-.05em{\sc i\kern-.025em b}\kern-.08em
    T\kern-.1667em\lower.7ex\hbox{E}\kern-.125emX}}
\begin{document}

\title{SEG:Seeds-Enhanced Iterative Refinement Graph Neural Network for Entity Alignment\\
}

\author{
	\IEEEauthorblockN{Wei Ai}
	\IEEEauthorblockA{\textit{College of Computer and Mathematics} \\
		\textit{Central South University of Forestry and Technology}\\
		ChangSha, China \\
		weiai@csuft.edu.cn}
	\and
	\IEEEauthorblockN{Yinghui Gao}
	\IEEEauthorblockA{\textit{College of Computer and Mathematics} \\
		\textit{Central South University of Forestry and Technology}\\
		ChangSha, China \\
		20221200518@csuft.edu.cn}
	\and
	\IEEEauthorblockN{\hspace{5em}Jianbin Li}
		\IEEEauthorblockA{\textit{\hspace{4em}College of Computer and Mathematics}\\
		\textit{\hspace{4em}Central South University of Forestry and Technology}\\
		\hspace{4em}ChangSha, China \\
		\hspace{4em}jianbinli@csuft.edu.cn}
	\and
	\IEEEauthorblockN{Jiayi Du}
	\IEEEauthorblockA{\textit{College of Computer and Mathematics} \\
		\textit{Central South University of Forestry and Technology}\\
		ChangSha, China \\
		dujiayi@csuft.edu.cn}
	\and
    \IEEEauthorblockN{\hspace{5em}Tao Meng{*}}
	\IEEEauthorblockA{\textit{\hspace{4em}College of Computer and Mathematics} \\
		\textit{\hspace{4em}Central South University of Forestry and Technology}\\
		\hspace{4em}ChangSha, China \\
		\hspace{4em}mengtao@hun.edu.cn}
	\and
	 \IEEEauthorblockN{\hspace{5em}Yuntao Shou}
	\IEEEauthorblockA{\textit{\hspace{4em}College of Computer and Mathematics} \\
	\textit{\hspace{4em}Central South University of Forestry and Technology}\\
	\hspace{4em}ChangSha, China \\
	\hspace{4em}shouyuntao@stu.xjtu.edu.cn}
	\and
	\IEEEauthorblockN{\hspace{2em}Keqin Li}
	\IEEEauthorblockA{\hspace{3em}Department of Computer Science \\
		\hspace{3em}State University of New York\\
		\hspace{3em}New Paltz, New York 12561, USA \\
		\hspace{4em}lik@newpaltz.edu}
	\thanks{ This work is supported by National Natural Science Foundation of China (Grant No. 69189338, Grant No. 62372478), Excellent Young Scholars of Hunan Province of China (Grant No. 22B0275), Changsha Natural Science Foundation (Grant No. kq2202294), and program of Research on Local Community Structure Detection Algorithms in Complex Networks (Grant No. 2020YJ009) \par * is the corresponding author.}
}

\maketitle

\begin{abstract}
Entity alignment is crucial for merging knowledge across knowledge graphs, as it matches entities with identical semantics. The standard method matches these entities based on their embedding similarities using semi-supervised learning. However, diverse data sources lead to non-isomorphic neighborhood structures for aligned entities, complicating alignment, especially for less common and sparsely connected entities. This paper presents a soft label propagation framework that integrates multi-source data and iterative seed enhancement, addressing scalability challenges in handling extensive datasets where scale computing excels. The framework uses seeds for anchoring and selects optimal relationship pairs to create soft labels rich in neighborhood features and semantic relationship data. A bidirectional weighted joint loss function is implemented, which reduces the distance between positive samples and differentially processes negative samples, taking into account the non-isomorphic neighborhood structures. Our method outperforms existing semi-supervised approaches, as evidenced by superior results on multiple datasets, significantly improving the quality of entity alignment.
\end{abstract}


\section{INTRODUCTION}
\begin{figure}
	\centering
	\includegraphics[width=0.4 \textwidth]{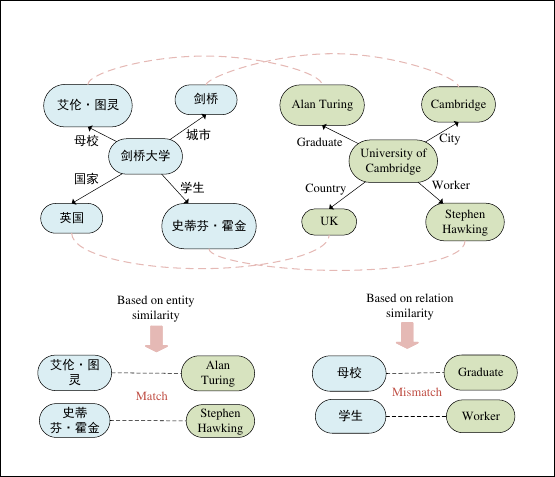}
	\caption{The difference in alignment for entities and relationships with non-isomorphic neighborhoods.}
	\label{fig:intro}
\end{figure}

\label{sec:Introduction}
Knowledge graph is a large-scale structured semantic knowledge base used to describe concepts in the physical world and their interrelationships, aggregate information and data, and link relationships on the Web into knowledge \cite{shou2022conversational, shou2025masked, shou2022object, shou2023comprehensive, shou2024adversarial, shou2023low, meng2024deep, shou2023adversarial}. Knowledge graphs contain rich information such as entity attributes, semantic relationships, and entity types. There are many large knowledge graphs, such as DBpedia \cite{auer2007dbpedia}and YAGO\cite{suchanek2007yago}, which have been widely used in natural language processing, information retrieval, and intelligent question answering. Due to the frequent inclusion of many entities and relationships in knowledge graphs, there is also a close connection with scale computing. However, knowledge graphs are inherently incomplete. If downstream applications need support, integrating various KGs for knowledge fusion effectively solves this problem. Researchers have proposed many entity alignment methods. People have gradually shifted from traditional  TransE~\cite{bordes2013translating, ai2023gcn, ai2024gcn, meng2024multi, shou2024contrastive, shou2024spegcl, shou2024efficient, ying2021prediction} methods to deep learning-based modeling of the graph structure of knowledge graphs. The methods based on deep learning~\cite{wang2018cross, shou2023graph, shou2023czl, meng2024masked, shou2024revisiting, ai2024edge, zhang2024multi} mainly follow the learning embedding-similarity calculation-greedy matching mode, which is to use seed entities to automatically extract implicit features of entities from different KGs using neural networks and then embed them into the same implicit space and calculate the similarity between different entity embeddings. However, in actual knowledge graphs, due to the heterogeneity of data sources, entities usually exhibit non-isomorphic neighborhood structures, and people have begun to realize that more is needed to consider entity embeddings for alignment. \cite{zhu2021relation, ai2023two, meng2024revisiting, ai2024mcsff, shou2023graphunet} considered the positive interaction between entity alignment and relationships in a semi-supervised context and mined relational information to assist in alignment. \cite{zhang2024relation, ai2024graph} considers the simultaneous roles of entities and relationships.
Although these scholars have attached varying degrees of importance to the importance of relationships in alignment, most methods still ignore two issues:\newline
\indent First, the difference between entities and relations is ignored. In the mainstream methods, embedding-similarity-matching, similarity methods can also be divided into two types: based on equivalent entities having similar attributes\cite{sun2018bootstrapping} and based on equivalent entities having similar adjacent entities. However, in these methods, the propagation of alignment information focuses more on the propagation from the similarity of entities to relations. Not only does it largely ignore the adequate auxiliary alignment information of the relational semantics itself, but there is also a problem of information loss. More importantly, it ignores the differences between entities and relations, which leads to inconsistent information. As shown in Fig \ref{fig:intro}, if only considering the propagation of similarity in entities, then Alan Turing and Alan Turing, Stephen Hawking, and Stephen Hawking can be fully aligned, but in relations, alma-mater, and graduate, student and worker cannot be aligned. GCN-based frameworks also use relations as weights\cite{cao2019multi} or information. However, noise is inevitably mixed in when the same model is used to train the representation of entities and relations jointly. The integration of relations into entity representation will cause the entity representation to be blurred and smooth.\newline
\indent Second, there is an obvious problem in current research: over-emphasis on reducing the distance between positive samples while ignoring the distance between negative samples, especially when facing pseudo-aligned entities with high similarity; there needs to be more differentiated treatment of these pseudo-entities. In cross-knowledge graph alignment, effectively dealing with the non-isomorphic neighborhood differences between positive and negative samples is a significant challenge. 
In order to overcome the limitations of existing methods, this paper proposes a soft label propagation framework based on multi-source fusion combined with an iterative process of seed enhancement. First, we conduct dual-angle modeling based on entities and relations and fuse the best-matching relationship pairs selected in both entity and relationship modes to generate soft labels. These soft labels contain rich entity neighbor features and integrate auxiliary information of relationship semantics, avoiding the problem of auxiliary information loss. Secondly, the soft labels fused by multi-source information propagate alignment information, and the topological structure and neighborhood differences are effectively captured. On this basis, a bidirectional adaptive weighted joint loss function is introduced, which not only shortens the distance between positive samples through commonalities but also widens the distance between high-similarity negative samples to varying degrees and more comprehensively considers the commonalities and differences in the alignment process. Some contributions are as follows: 

\begin{itemize}

\item
	Obtaining soft labels from the perspectives of entities and relations can reveal shared entity features and mine the rich information of the relationship semantics itself, thereby improving alignment accuracy.
	
\item
	The difference problem of non-isomorphic neighborhoods is solved by effectively shortening the distance between positive samples and using a weighted joint loss function to differentiate and distance negative samples.
	
\item
	Our method, tested on real datasets, significantly boosts model performance, proving highly feasible and valuable for entity alignment across knowledge graphs.

\end{itemize}

The remainder of this paper is organized as follows: Section \ref{sec:related_work} reviews related work, Section \ref{sec:proposed_method} provides a detailed description of the proposed method, Section \ref{sec:experrments} presents the experimental setup and results, and finally, Section \ref{sec:conclusion} offers a brief conclusion.

\begin{figure*}
	\centering
	\includegraphics[width=0.8 \textwidth]{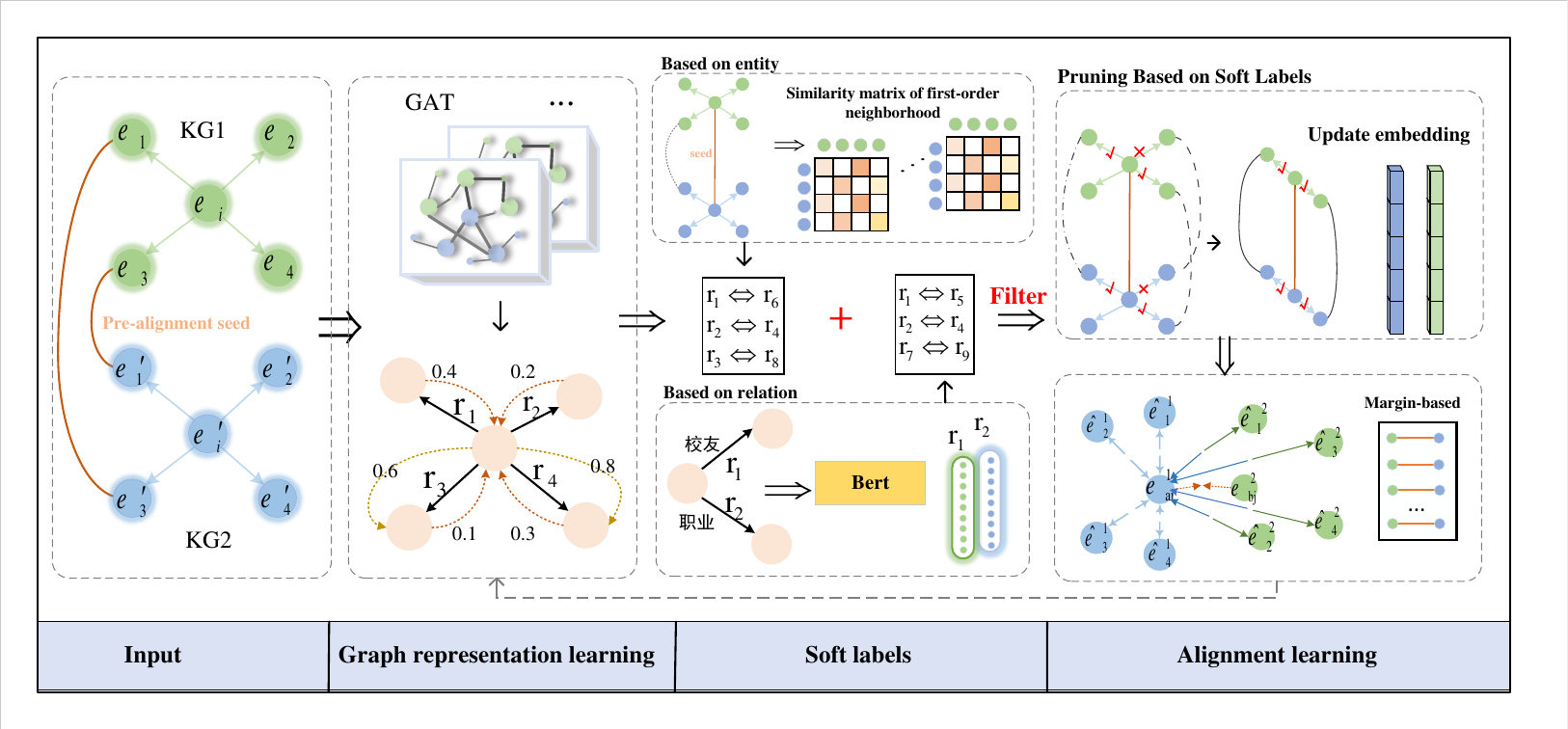}
	\caption{The entity alignment approach of SEG, as outlined in the framework diagram, encompasses four principal components: Graph representation learning, Soft Labels, Pruning Based on Soft Labels, and Bidirectional Weighting Mechanism.}
	\label{fig:framework}
\end{figure*}

\section{RELATED WORK}
\label{sec:related_work}
In this part, we examine existing scholarly contributions about entity alignment, encompassing conventional translation-based tactics as well as contemporary graph representational techniques that leverage Graph Convolutional Networks (GCNs)\cite{chen2019gated}\cite{chen2020citywide}.
\subsection{Conventional Approaches Utilizing Translation Techniques}
Translation-based entity alignment. By continuously adjusting the vector representation, the entity vector + the relation vector = the entity vector. In order to solve the shortcomings of TransE. MTransE\cite{chen2016multilingual} embeds entities and relations in different knowledge graphs into different vector spaces and maps them to a unified space through distance calibration and linear transformation strategies to achieve cross-knowledge graph alignment. BootEA\cite{sun2018bootstrapping} is a Bootstrap-based entity alignment method that converts entity alignment into a classification problem. It completes the alignment task by iteratively marking possible entity alignments as training data and finding the maximum possibility of alignment. 
\subsection{GCN-Driven Graph Representation Techniques
}
GNN-based entity alignment. GCN-align can embed entities from different KGs into a unified vector space. Embedding can be learned from the structure and attribute information of the entity. HMAN\cite{yang2019aligning} embeds learning entity representation from three perspectives: structural information, relationship features, and attribute characteristics.RD-GCN\cite{wu2019relation} uses the graph attention mechanism to encourage dual relations between the interaction graph and the original graph and then feeds the vertex representation in the original graph into the GCN layer with highway neural network gating to capture the structural information of the neighbors. 
\subsection{Semi-supervised entity alignment}

\section{PROPOSED METHOD}
\label{sec:proposed_method}
We introduce the SEG methodology, whose structural outline is depicted in Fig \ref{fig:framework}. Entity encoding and alignment candidate generation.Soft label screening: The soft labels of the two modalities are fused, and the screening strategy is used to obtain the final accurate soft labels.Pruning when performing neighborhood aggregation based on soft labels: Propagate the alignment relationship in the soft labels to the neighborhood structure of the candidate alignment.Bidirectional weighted joint mechanism: Use a weighted method to differentiate negative samples to guide positive and negative sample learning between two knowledge graphs.Bidirectional iterative collaborative training.\newline
\indent In this study, we consider the scenario where two distinct knowledge graphs are provided as inputs. A knowledge graph is structured as $KG = (E, R, T)$, where $E$ denotes the set of entities, $R$ represents the set of relations, and $T$ is a collection of triples defined as $\{(e_1, r, e_2)|e_1, e_2 \in E, r \in R\}$. The source knowledge graph is denoted by $kG_1=(E^1, R^1, T^1)$; similarly, the target knowledge graph is denoted by $KG_2=(E^2, R^2, T^2)$. The set of seed alignment entities is given by $Align_{seed}=\{(u,v)|u \in E^1,v \in E^2,u \leftrightarrow v\}$, indicating that entity $u$ from $KG_1$ and entity $v$ from $KG_2$ are identical, with $ \leftrightarrow $ signifying equivalence. To initiate the process, we employ a pre-trained model to assign initial embeddings to the entities.

\subsection{Graph representation learning}
The Graph Attention Network serves as a feature extraction mechanism, introducing an innovative neural network framework for processing data structured as a graph. This architecture comprises an input layer, an intermediary hidden layer, and a final output layer. Within each hidden layer, the representation of a node is determined by its interaction with the surrounding nodes' information, necessitating ongoing updates to the hidden layer's state. The computation is based on the following formula:

\begin{equation}
H_{e_i}^{(l+1)} = H_{e_i}^{(l)} + d^{(l)} \cdot \sigma(\sum_{H_{e_j} \in N_i} a_{ij}^{(l)} H_{e_j}^{(l)}),
\end{equation}

where $H_{e_i}^{(l+1)}$ and $H_{e_i}^{(l)}$ represent the embedding of the entity at the $l$ layer, $l$ represents the number of layers, and $d^{(l)}$ represents the weight parameter of the  $l$ layer,$\sigma$ represents the ReLU activation function.

\begin{equation}
	a_{ij} = \frac{\exp (\varepsilon S_{ij})}{\sum_{H_{e_k} \in N(H_{e_i})} \exp (\varepsilon S_{ik})}
\end{equation}

\begin{equation}
	S_{ij} = \sum_{(H_{e_i},R_t,H_{e_j}) \in T_{ij}} a^T ([H_{e_i}||H_{e_j}] \odot R_t)
\end{equation}

where $T_{ij}$ represents the set of relation triplets with $e_i$  as the head entity and $e_j$ as the tail entity, $\alpha$ is the attention parameter, $ || $ are vector concatenation, $\odot$ element multiplication and LeakyReLU respectively function.

\vspace{-4mm}
\begin{equation}
	\begin{split}
		R_t = \sigma \left[ \frac{\sum_{H_{e_i} \in T_t} b^T H_{e_i}}{|T_t|} \bigg\| \frac{\sum_{H_{e_j} \in W_t} b^T H_{e_j}}{|W_t|} \right],
	\end{split}
\end{equation}

Where $T_t$ and $W_t$ represent the size of the head entity set and the tail entity set connected by $R_t$, respectively, $b$ is the attention parameter.

\begin{equation}
	H_{e_i} = W^n \cdot n(H_{e_i}) + b^n.
\end{equation}

$w$ and $b$ are trainable weight matrices, and their output matrices are represented as $H_{e_i}$.

\indent Alignment candidates.
After obtaining the embedding representation of the entity through $GAT $, the embedding representation of the two heterogeneous knowledge graphs is obtained respectively. $ E^1 = \{e_1^1, e_2^1, e_3^1 \ldots, e_i^1\}, E^1 \in KG_1, E^2 = \{e_1^2, e_2^2 e_3^2, \ldots, e_j^2\}, E^2 \in KG_2, $, and then the candidate entity pairs need to be matched, that is, $ e_i^1 \leftrightarrow e_j^2 $, cosine similarity is used to evaluate the similarity between entities. The similarity calculation results of all embeddings in $E^1$ and $E^1$ can be calculated according to the cosine similarity, and the calculation results are saved in the similarity matrix cosine\_distances. The calculation formula is as follows:

\begin{equation}
	Sim_{(e_i, e_j)} = \sum_{i \in KG_1, j \in KG_2}^n \frac{e_i \cdot e_j}{\|e_i\| \cdot \|e_j\|},
\end{equation}

$Sim \in [-1, 1]$, Only the similarity in the matrix $>=$ 0.95 is selected as the Candidate pre-selected set. However, these candidate sets only represent possible matching entity pairs, and there may even be one-to-many situations that cannot guarantee the accuracy of the final alignment.

\subsection{Soft Labels}
\indent Based on Entity propagate similarity. taking the aligned seeds of $KG_1$ and $KG_2$ as anchors, for any seed pair $(e_a^1, e_b^2) \in Align_{entity}$, we construct the neighbourhood node sets of $e_a^1$ and $e_b^2$, ${n(\bar{e}_i^1)|\bar{e}_i^1 \in N(e_a^1)}$ and ${n(\bar{e}_j^2)|\bar{e}_j^2 \in N(e_b^2)}$, $(e_a^1, r_{a1}, \bar{e}_1^1)$ or $(\bar{e}_1^1, r_{a1}, e_a^1)$. $n(\bar{e}_i^1)$ represents the $i$-th neighbor node of $e_a^1$, and $r_{a1}$ represents the connection between the two. $\bar{e}_j^2$ represents the $j$-th neighbor node of $e_b^2$, $(e_b^2, r_{b1}, \bar{e}_1^2)$, $(\bar{e}_1^2, r_{b1}, e_b^2)$, $r_{b1}$ is the same. Here, the above cosine similarity is also used to calculate the similarity of the nodes $(e_a^1, e_b^2)$ in the neighborhood of the seed pair. In order to ensure that the relationship pairs screened by the neighborhood candidate alignment set are highly accurate and convenient for subsequent training, two standards are set here: First, the neighborhood alignment nodes need to meet the set $Sim_e$ similarity threshold and the $Match_e$ matching number threshold of the derived relationship. Second, we must ensure that the neighborhood nodes are aligned one-to-one. Only the neighborhood pairs with the highest similarity are selected when encountering a one-to-many or many-to-one situation.
\begin{equation}
	Sum_1 = \{Sim_e >= 0.98\} + \{Match_e >= 10\},
\end{equation}
$Sum_1$ represents the soft label set screened based on the entity embedding model, $Sim_e$ represents the cosine similarity of the neighborhood node pair, and match is the number of repetitions of the relationship between the entities. For example: $(e_a^1, e_b^2) \to (e_a^1, r_{a1}, \bar{e}_1^1), \quad (e_b^2, r_{b1}, \bar{e}_1^2) \to \text{Sim}(\bar{e}_1^1, \bar{e}_1^2) >= 0.98$ and $\text{Match}(r_{a1}, r_{b1}) >=10$,that is, keep this relationship pair in the set $Sum_1$.

\indent Based on relation propagate similarity. In this section, the Bert model is introduced to encode the relations of the KG graph, and the semantic features of the relations are fully captured using the name and description of the relations. 
\begin{equation}
	H_r = \text{Bert}(RD),
\end{equation}
RD: Relation Description. Similarly, setting the cosine similarity and matching threshold also requires one-to-one matching.
\begin{equation}
	Sum_2 = \{Sim_r >= 0.98\} + \{Match_r >= 600\},
\end{equation}

$Sum_2$ represents the number of soft labels selected based on the relation embedding model, $Sim_r$ represents the cosine similarity of the relation pair, and $Match_r$ is the number of times this relation pair appears repeatedly. $\text{Sum} = \text{Sum}_1 + \text{Sum}_2, \quad \text{Order}(\text{Sum}_2) > \text{Order}({Sum}_1)$, $ { Sum }_2 $has a higher priority than $ { Sum }_2 $.

\subsection{Pruning Based on Soft Labels}
Soft labels enhance alignment precision by guiding neighborhood matching and are selectively retained based on match quality. GCN integrates this neighborhood data, while Highway networks combine entity names with neighborhood information to refine the final embeddings.

\subsection{Bidirectional Weighting Mechanism}
Inspired by contrastive learning, we have engineered a bidirectional weighting mechanism based on similarity aimed at optimizing the weight distribution of samples during the entity alignment process. This mechanism adjusts the weights of negative samples differentially to achieve a precise approximation of positive samples and effectively optimize negative samples. Positive samples are actual alignments $Align_{entity}$, while negative samples are selected from the top 50 nodes with the highest similarity to the positive samples in either $KG_1$ or $KG_2$. Notably, due to the most significant risk of misalignment, the node with the highest similarity is assigned the highest weight penalty.

\begin{equation}
	Loss_w = \sum_{i=1}^P \sum_{j=1}^N \beta \exp\left(-\gamma\frac{j-1}{K-1}\right) \cdot \mathcal{D}_{ij},
\end{equation}

\begin{table*}[htbp]
	\renewcommand\arraystretch{1.4}
	\caption{Ablation Study and Comparative and Proprietary Method Performance on DBP15K Language Datasets using Hit@1, Hit@5, and MRR Metrics.}
	\begin{center}
		\setlength{\tabcolsep}{2.8mm}{
			\begin{tabular}{ccccccccccccc}
				\hline
				\multirow{2}{*}{\textbf{Method}} &&\multicolumn{3}{c}{\textbf{DBP$_{JA-EN}$}}&&\multicolumn{3}{c}{\textbf{DBP$_{FR-EN}$}}&&\multicolumn{3}{c}{\textbf{DBP$_{ZH-EN}$}} \\
				\cline{3-5} \cline{7-9} \cline{11-13}
				~ && \textbf{\textit{Hit@1}}& \textbf{\textit{Hit@5}} & \textbf{\textit{MRR}} && \textbf{\textit{Hit@1}}& \textbf{\textit{Hit@5}} & \textbf{\textit{MRR}} && \textbf{\textit{Hit@1}}& \textbf{\textit{Hit@5}} & \textbf{\textit{MRR}}\\
				\hline
				SEA    &  & 29.12 & 55.64 & 0.416 & & 32.64 & 60.28 & 0.455 & & 34.56 & 59.54 & 0.464 \\
				
				BootEA   &  & 52.71 & 71.89 & 0.616 & & 57.61 & 77.27 & 0.666 & & 55.45 & 73.72 & 0.639 \\
				\hline
				RAGA &  & 79.29 & 89.12 & 0.838 & & 85.27 & 93.17 & 0.889 & & 68.72 & 82.55 & 0.750 \\
				
				RANM     &  & 90.56 & 94.30 & 0.923 & & 90.96 & 94.74 & 0.927 & & 77.64 & 85.66 & 0.813 \\
				
				SEG             &  & \textbf{91.56} & \textbf{95.2} & \textbf{0.933} & & \textbf{92.26} & \textbf{95.44} & \textbf{0.937} & & \textbf{78.55} & \textbf{86.53} & \textbf{0.822} \\
				\hline
				SEG( w / 0  BWM)    &  & 91.0 & 94.07 & 0.924 & & 92.18 & 95.47 & 0.937 & & 78.2 & 86.16 & 0.819 \\
				SEG( w / 0 SL)    &  & 91.53 & 95.10 & 0.932 & & 91.77 & 95.8 & 0.936 & & 78.04 & 85.89 & 0.817 \\
				\hline
			\end{tabular}
		}
		\label{tab3}
	\end{center}
\end{table*}

P represents the number of positive samples, while N denotes the number of negative samples corresponding to each positive sample (n\_negatives\_per\_positive).$D_{ij}$ represents the distance of features. The feature vector of the j-th negative sample corresponds to the i-th positive sample, where $j \in \{1,\ldots, K\}$,$\gamma$ and $\beta$ are hyperparameters.

\textbf{{Alignment learning}}. In this paper, a margin-based loss function is used. Moreover, The entity alignment training optimization in this paper adopts a semi-supervised learning method and expands the training data through an iterative strategy to reduce the dependence on manually labelled seeds and the consumption of workforce and material resources. 

\begin{align}
	Loss_m &= \sum_{(e^1,e^2)\in P,(e^{1'},e^{2'})\in N} \left[D(e^1,e^2) + \gamma - D\left(e^1,e^{2'}\right)\right]_+ \notag \\
	&+ \sum_{(e^1,e^2)\in P,(e^{1'},e^{2'})\in N} \left[D(e^1,e^2) + \gamma - D\left(e^{1'},e^2\right)\right]_+
\end{align}

\begin{equation}
	Loss = Loss_w + Loss_m
\end{equation}

$[x]_+ = \max\{0, x\},\quad(e^1, e^2) \in P$ represents the alignment seed positive sample set, and $e^1\in E^1,\quad E^1 \in KG_1,\quad e^2$ is the same,$\quad(e^{1'}, e^{2'})\in N$ represents the alignment seed negative sample set,$\quad D(e^1,e^2) = \|e^1 - e^2\|_{L_1}$represents the $L_1$ function between two vectors,$\quad\gamma > 0$ is a marginal hyperparameter.

\section{EXPERIMENTS}
\label{sec:experrments}
This section presents the experimental settings, including experimental datasets, evaluation metrics, model variants, and experimental implementation. The model in this paper is tested on the DBP15K datasets to study further the essential components of this model and their contributions.
\subsection{Datasets}
DBP-15K is built on DBpedia and includes three cross-language DBpedia datasets: ZH-EN (Chinese-English), JA-EN (Japanese-English), and FR-EN (French-English).

\begin{table}[htbp]
	\renewcommand\arraystretch{1.2}
	\caption{Statistic dates}
	\begin{center}
		\setlength{\tabcolsep}{1.3mm}{
			\begin{tabular}{c|c|c|c}
				\hline
				\multirow{2}{*}{\textbf{Datasets}} & \multicolumn{3}{|c}{\textbf{Graph basic information statistics}} \\
				\cline{2-4}
				~ & \textbf{\textit{Entity}} & \textbf{\textit{Relations}} & \textbf{\textit{Rel.triple}} \\
				\hline
				\multirow{2}{*}{JA-EN}  & 19,814   & 1,299 & 77,214 \\
				\cline{2-4}
				~                      & 19,780   & 1,153 & 93,484 \\
				\hline
				\multirow{2}{*}{FR-EN}  & 19,661   & 903   & 105,998 \\
				\cline{2-4}
				~                      & 19,663  & 1,208 & 115,722 \\
				\hline
				\multirow{2}{*}{ZH-EN}  & 19,661   & 903 & 105,998 \\
				\cline{2-4}
				~                      & 19,993   & 1,208 & 115,722 \\
				\hline
				\multicolumn{3}{l}{$^{\prime}$Abbreviated form.}
			\end{tabular}
		}
		\label{tab1}
	\end{center}
\end{table}

\vspace{-2.5mm}

\subsection{Baseline}
To assess the performance of our model, we conducted comparisons against current leading entity alignment techniques. Broadly, these can be categorized as follows:
\indent Traditional Entity Alignment Methods. SEA proposes a semi-supervised entity alignment method that uses labeled entities and rich unlabeled entity information for alignment. BootEA proposes a Bootstrap-based entity alignment method. Relation-based Entity Alignment Methods. RAGA proposed a framework for capturing entity and relation interactions based on relation-aware graph attention networks. RANM\cite{cai2022semi} also demonstrates exceptional performance in leveraging relational assistance for entity alignment.

\subsection{Evaluation setting}
To ensure that the model remains optimal, five-fold cross-validation based on the training set (20\%), validation set (10\%), and test set (70\%) was used, so the reported performance is the average of five independent training runs, and the training/validation/test datasets are shuffled in each round. The number of GAT layers is 2, the similarity threshold of entity embedding is 0.95, and the maximum number of neighbor nodes for alignment seeds is 982. The final training parameters are selected between the best performance period on the validation set (based on "Hits@1") and the last period (up to 1500 periods). The experiments were run on NVIDIA 4090 16 GB.

\subsection{Ablation Studies}
Our model exceeds baselines, as detailed in the table. Ablation studies adjusted parameters like GNN layers and learning rate, showing that the soft labels module enhanced DBP-15K accuracy by 0.3\%-1\%, especially in the ZH-EN subset with a 1\% gain. Additionally, the Bidirectional Weighting module raised DBP-15K accuracy by 0.2\%-0.9\%, contributing to a consistent 0.7\%-1.3\% overall improvement across all subsets.

\subsection{Robustness Analysis}
In assessing our model's robustness, we tested the impact of GNN layer count and learning rate. While more GNN layers slightly lowered entity alignment accuracy but kept overall performance stable, higher learning rates (0.01, 0.005, 0.001) significantly reduced performance due to training instability and convergence challenges.

\section{CONCLUSION}
\label{sec:conclusion}
This paper presents the SEG framework for cross-lingual entity alignment in knowledge graphs, which enhances entity information through a multi-source fusion mechanism and utilizes semantic auxiliary information from relations to assist in entity alignment. Additionally, it employs a bidirectional weighting mechanism to improve the model's differential processing of negative samples, complemented by a semi-supervised iterative mechanism that continually expands and strengthens alignment seeds. The robustness of this method has been validated on datasets, and future efforts will concentrate on refining relation representation and entity enhancement techniques to enhance alignment accuracy.

\label{sec:references}
\bibliographystyle{IEEEtran}
\bibliography{refer}

\end{document}